# Graphical models for preference and utility

**Fahiem Bacchus**
Dept. of Computer Science
University of Waterloo
Waterloo, Ontario
Canada, N2L 3G1
fbacchus@logos.uwaterloo.ca

**Adam Grove**
NEC Research Institute
4 Independence Way
Princeton NJ 08540, USA
grove@research.nj.nec.com

## Abstract

Probabilistic independence can dramatically simplify the task of eliciting, representing, and computing with probabilities in large domains. A key technique in achieving these benefits is the idea of graphical modeling. We survey existing notions of independence for *utility* functions in a multi-attribute space, and suggest that these can be used to achieve similar advantages.

Our new results concern conditional additive independence, which we show always has a perfect representation as separation in an undirected graph (a Markov network). Conditional additive independencies entail a particular functional form for the utility function that is analogous to a product decomposition of a probability function, and confers analogous benefits. This functional form has been utilized in the Bayesian network and influence diagram literature, but generally without an explanation in terms of independence. The functional form yields a decomposition of the utility function that can greatly speed up expected utility calculations, particularly when the utility graph has a similar topology to the probabilistic network being used.

## 1 Introduction

Work over the past decade in artificial intelligence concerning probabilities has been extremely successful. Charges of *epistemological inadequacy* [MH69] have become much easier to rebut since the advent of Bayesian networks and similar techniques.

But probabilities are not an end in themselves. Their most important purpose is in classical decision theory, as part of the maximum expected utility paradigm (see, e.g., [Fre88, GS88, Sav54]). This leads to the concern that other parts of decision theory might not be keeping pace with the new developments in probabilistic modeling. In particular, we are interested in the problem of representing, and reasoning about, *utility* and *preference*.

There are a number of important questions in this area, and here we report on our early observations and results and make a few conjectures about promising directions for future research. The approach we take is based on the idea of drawing a close analogy between probabilities and utilities. It is clear that some of the issues are similar:

- The "too many numbers!" criticism of probability theory can apply to utilities as well. The number of possible worlds grows exponentially in the number of properties (i.e., attributes or variables) used to describe them. In principle, each world might require an independently assigned utility. Of course, we may be lucky enough that a world's utility depends on just a few attributes. In this case many different possible worlds will have the same utility and we will have far fewer distinct values to deal with. But we cannot rely on this in general.

- Probabilities can be difficult to elicit and to compute with. For this reason, there have been many attempts to deal with uncertainty in a way that avoids probability, even though many (although not all) of these approaches are *ad hoc* and lack any solid foundation.

  Correspondingly, there are several common ways to describe how one wants a complex system to behave short of actually giving utilities: e.g., one can list preferred *goals*, impose hard or soft *constraints* on behavior, and so on.

Graphical models, such as Bayesian networks, address many of the perceived problems of probability in a fairly successful fashion (see, e.g., [Pea88, SP90] for an introduction to this area). The key trick, of course, is probabilistic independence. Independence can vastly reduce dimensionality, in the sense of the number of independent parameters we must discover. We can make many judgments of independence based on a qualitative (typically causal) understanding of the domain, and afterwards elicit or learn the remaining conditional probabilities. But reduction of dimensionality is not the whole story: probabilistic independence lends itself to graphical representations that greatly aid intuition and support relatively efficient computational techniques.

Can a similar story be told for utilities? This is plausible because utility, like probability, is often highly structured.

Furthermore, much (but not all) of this structure can be described in terms of *independence*, using one or all of several independence concepts that already exist in the literature. We begin this paper, in Section 2, with background material including a brief survey of some existing independence concepts in utility theory. We note that these concepts are part of a well established field known as *multiple-objective decision theory*, for which [KR76] is an excellent reference.

So far as we are aware, the relevance of the results in this field for artificial intelligence is a largely unexplored topic (although [DW91, DSW91, DW94, DW92] are exceptions to this, and there is a growing collection of work concerned with other aspects of utility such as [Bou94, TP94]). Our main results in this paper are in Section 3 and concern *conditional additive independence* (CA-independence), whose definition is reviewed in Section 2. This concept seems to strike a good balance between being too weak (thus leading to few useful conclusions) and too stringent (thus being infrequently applicable).

Our first result in Section 3 shows that this notion has a precise representation as separation in undirected graphs. This result is closely related to the theory of graphoids and Markov fields; see [Pea88] for a description of these notions.

A utility function satisfies a CA-independence if and only if it can be written in a particular functional form. This functional form is an additive decomposition of the utility function, that is analogous to a product decomposition of a probability function and leads to a similar reduction of dimensionality. Our results show how this functional form can be read directly from the graph that represents the CA-independencies of the utility function.

As we briefly discuss in Section 3.1, an additive utility decomposition can simplify expected utility and related computations. Tatman and Shachter [ST90] show one way to take advantage of this phenomenon in influence diagram computation. If we are given a probabilistic network with a topology that is "similar" to our utility graph, the potential for computational speedup would appear to be especially great. This is implicitly reflected in Jensen et al.'s work [JJD94], which gives another technique for evaluating influence diagrams.

Both [JJD94] and [ST90] take an additive utility function as their starting point. They do not address the question of where the decomposition comes from. In that sense the results of this paper can be seen to be complementary to these works. We discuss the notions of utility independence that allow such decompositions of the utility function, and make a start at providing some graphical modeling tools for dealing with these notions of independence. The final part of this paper, Section 4, briefly mentions some topics that we believe are promising directions for future work.

## 2   Preliminaries

Decision theory is useful in a setting where the system (the world) may end up in one of several possible states. If we have some control (via our actions) as to which state obtains, we need to know how to choose the best action. As is well known, there are several axiomatizations of "rational" decision making that lead to the maximum expected utility criterion (see, e.g., [Fre88, GS88, Sav54]). This says that we should attempt to maximize the sum, over all states, of the product of the *probability* and the *utility* of each state. So if the probability distribution depends on the action we take, this criterion can determine the best action. We begin with a very quick review of the relevant concepts, mainly to set up the necessary notation. This review is based on the following sources [Fis82, Fre88, KR76, KLST71].

If there are $N$ states a probability distribution and utility function have $N-1$ and $N-2$ independent parameters, respectively.[1] Unfortunately $N$ is often *very* large, exponential in the number of attributes or variables we use to describe a state. Thus it is important that the utility function possess some structure so as to simplify the tasks of elicitation, representation, and computation. This is of course exactly what graphical models based on independence try to achieve for probabilities.

As in the probabilistic network literature, our first assumption is that the set of states can be represented as a product space over some set of attributes or variables.

**Notation:** Throughout this paper, we assume that $V = \{v_1, \ldots, v_n\}$ is a fixed set of $n$ *variables*. Each variable $v$ has a domain $d_v$ of two or more elements.[2] We will generally use lower case letters to denote variables and upper case letters to denote sets of variables. (Note that this is somewhat nonstandard.) Where necessary, Greek letters will denote values for particular variables.

The set of states $S$ consists of the set of points in the product space $\prod_{i=1}^n d_{v_i}$. Each $s \in S$ is thus a vector of $n$ values, one value for every variable. Clearly the size of $S$ is exponential in the number of variables.

If $X \subseteq V$ then $f(X)$ stands for some real valued function all of whose arguments are in $X$, i.e.,

$$f(X) : \prod_{v \in X} d_v \longrightarrow \mathbb{R}$$

The general form of a utility function is $u(V)$, which can thus require exponentially many independent utility assessments.

A utility function $u$ induces a *preference ordering* $\succeq_u$ on *lotteries*[3] (probability distributions) over $S$ as follows:

$$p_1 \succeq_u p_2 \quad \textit{iff} \quad \sum_{s \in S} p_1(s) u(s) \geq \sum_{s \in S} p_2(s) u(s),$$

---

[1] Utility theory is invariant with respect to affine transformations of the utility function, which is why only $N-2$ independent utilities need be found.

[2] Everything we say applies for infinite domains as well. Although we implicitly assume that domains are finite in parts of the following, this is for notational and conceptual simplicity only.

[3] "Lotteries" is one of the traditional terms. It can be misleading in that it tends to imply that the probabilistic structure arises from explicit randomization or "objective" randomness. This may be the case, but on the other hand the probabilities can also be an entirely subjective measure of uncertainty.

where $p_1$ and $p_2$ are two distributions over $S$. That is, we prefer $p_1$ to $p_2$ if $p_1$ induces greater expected utility. Thus utility serves to characterize not only the agent's values but also its attitudes towards risk: it ranks probabilistic gambles between various outcomes.

In the development of decision theory, it is natural to take the preference relation as primitive. Any relation satisfying fairly weak rationality conditions (which we don't repeat here, but see, e.g., [Sav54, Fis82, Fre88]) corresponds to some utility function exactly as above (that is, furthermore, unique up to affine transformations). This exact correspondence between preference and utility is one of the fundamental theorems of decision theory. In the following, whenever we talk about a preference over $V$ we mean a preference over lotteries over $S = \prod_{v \in V} d_v$ satisfying the standard rationality postulates.

The first definition of independence we consider is *utility independence*. Intuitively, a set of attributes $X$ is utility independent of everything else, if when we hold everything else fixed (i.e., the values of attributes $V-X$), the induced preference structure over $X$ does not depend on the particular values that $V-X$ are fixed to. Given utility independence we can assert preferences over (lotteries on) $X$ that hold *ceteris paribus*—i.e., all else being equal.

**Definition 2.1**: Consider preference $\succeq$ over $V$, $X \subset V$, $Y = V - X$. Let $\tilde{\gamma}$ be any particular element of $\prod_{v \in Y} d_v$. That is, $\tilde{\gamma}$ is a particular assignment of values to the variables in $Y$. Every probability distribution $p$ over $\prod_{v \in X} d_v$ corresponds to a distribution $p^*$ on $S = \prod_{v \in V} d_v$ such that $p^*$'s marginal on $X$ is $p$ and $p^*$'s marginal on $Y$ gives probability 1 to $\tilde{\gamma}$. We define the *conditional preference over $X$ given $\tilde{\gamma}$*, $\succeq_{\tilde{\gamma}}$, to be the preference ordering such that

$$p \succeq_{\tilde{\gamma}} q \quad \text{iff} \quad p^* \succeq q^*,$$

where $p$ and $q$ are any two distributions over $\prod_{v \in X} d_v$. ∎

**Definition 2.2:** The set of attributes $X$ is *utility independent* of $V-X$ when conditional preferences for lotteries on $X$ do not depend on the particular value given to $V-X$. That is,

$$\left( \forall \gamma, \gamma' \in \prod_{v \in V-X} d_v \right) p \succeq_\gamma q \text{ iff } p \succeq_{\gamma'} q,$$

where $p$ and $q$ are any two distributions over $\prod_{v \in X} d_v$. ∎

Utility independence fails, for instance, if one has a preference reversal between two mixtures of the attributes $X$, when some attribute in $V-X$ is changed. Judgments of utility independence would appear to be fairly natural and common; see [KR76] for a very extensive discussion. They are, at heart, judgments about *relevance* and people seem to be fairly good at this in general.

**Example 2.3:** Say that there are only two attributes *health*, with values $H$ and $\overline{H}$ (healthy and not healthy), and *wealth* with values $W$ and $\overline{W}$ (wealthy and not wealthy). If the agent's utility function $u$ is defined as $u(HW) = 5$, $u(H\overline{W}) = 2$, $u(\overline{H}W) = 1$, and $u(\overline{H}\overline{W}) = 0$, then it can be seen that for the agent *health* is utility independent of *wealth* and *wealth* is utility independent of *health*. Intuitively, no matter what the agent's wealth is fixed to, it will always prefer gambles that yield $H$ with higher probability. That is, the agent's preference for being healthy is the same no matter if the agent is wealthy or not. The same can be said about its attitude towards being wealthy. ∎

Utility independence is known to have several strong implications. We list a few, using [KR76] as our source. First, utility independence is equivalent to the existence of a utility function with a special functional form:

**Proposition 2.4:** $X$ *is utility independent of its complement in a preference structure $\succeq$ if and only if $\succeq$ corresponds to some utility function of the form:*

$$u_{\succeq}(V) = f(V-X) + g(V-X)h(X)$$

*where $g$ is positive.*[4]

Thus we must assess three functions, but each has fewer than $|V|$ arguments. This may mean that there are far fewer independent numbers to learn and to store. Most of the interest in utility independence in standard decision theory concerns the case of *mutual utility independence* where every subset of variables is independent of its complement:

**Proposition 2.5:** *Every subset of variables is independent of its complement in $\succeq$ if and only if there exists $n$ functions $f_i(v_i)$ (i.e., each $f_i$ depends on a single variable), such that either*

$$u_{\succeq}(X) = \prod_{i=1}^{n} f_i(v_i) + c$$

*for some constant $c$, or*

$$u_{\succeq}(X) = \sum_{i=1}^{n} f_i(v_i).$$[5]

This is an extremely strong conclusion, allowing enormous simplification. The precondition of the theorem might seem to require $O(2^n)$ utility independence conditions, but since utility independence satisfies various closure properties we do not need this many. There are in fact several sets of $n$ independencies that suffice; see [KR76]. However, the $n$ assertions that each attribute individually is independent of the rest are not sufficient. In this case, the result is weaker:

**Proposition 2.6:** *If every variable is utility independent of the rest there is a function $f_i(v_i)$ for each variable, such that $u_{\succeq}(V)$ is a multilinear combination of the $f_i$'s.*

Thus we must assess $n$ functions as well as (potentially exponentially many) constants to capture the interactions among the $f_i$'s. This may still represent a net gain. We suggest in Section 4 that this case might be important for artificial intelligence, and deserves future work.

---

[4] It is also clearly possible to arrange $f$ and $g$ so that $h(X) = u_{\succeq}(X, \tilde{\gamma})$ where $\tilde{\gamma}$ is an arbitrary fixed assignment to $V-X$. The function $u_{\succeq}(X, \tilde{\gamma})$ is sometimes called a *conditional utility function*.

[5] It is more usual to express the $f_i$ in terms of conditional conditional utility functions and multiplicative constants. This representation is easy to derive, or see [KR76].

A much stronger form of independence is *additive independence*. This can be defined in several ways, but the most useful for us is:

**Definition 2.7:** Let $Z_1, ..., Z_k$ be a partition of $V$. $Z_1, ..., Z_k$ is additively independent (for $\succeq$) if, for any probability distributions $p_1$ and $p_2$ that have the same marginals on $Z_i$ for all $i$, $p_1$ and $p_2$ are indifferent under $\succeq$, i.e., $p_1 \succeq p_2$ and $p_2 \succeq p_1$. ∎

In other words, one's preference only depends on the marginal probabilities of the given sets of variables, and not on any correlation between them.

**Example 2.8:** Consider the utility function given in Example 2.3 involving *health* and *wealth*. As the previous example pointed out, *health* was utility independent of *wealth*. However *health* is *not* additively independent of *wealth*. Consider the two probability functions $p_1$ and $p_2$, where $p_1(HW) = p_1(H\overline{W}) = p_1(\overline{H}W) = p_1(\overline{HW}) = 1/4$, and $p_2(H\overline{W}) = p_2(\overline{H}W) = 0$, $p_2(HW) = p_2(\overline{HW}) = 1/2$. We have $p_1(H) = p_2(H) = 1/2$ and $p_1(W) = p_2(W) = 1/2$. That is, $p_1$ and $p_2$ have identical marginals over *health* and *wealth*. Yet the expected utility under $p_1$ is 2, while the expected utility under $p_2$ is 5/2. This shows that there exists two distributions with the same marginals that are not indifferent under the given utility function. That is, *health* and *wealth* are not additively independent.

Intuitively, the agent prefers being both healthy and wealthy more than might be suggested by considering the two attributes separately. It thus displays a preference for probability distributions in which health and wealth are positively correlated. ∎

**Proposition 2.9:** $Z_1, ..., Z_k$ are additively independent for $\succeq$ iff $u_\succeq$ can be written as

$$u_\succeq(V) = \sum_{i=1}^{k} f(Z_i)$$

*for some functions $f_i$.*

Naturally, the most interesting case is where all variables are additively independent separately, so that we only need to find one single-argument function for each variable. In the rest of the paper, we will be interested in additive independence for a partition of $V$ into two parts, $V = X \cup Y$, unless we say otherwise. It would seem reasonable that these are easier to reason with than independence assertions about arbitrary partitions.

Conditional versions of both additive and utility independence can be defined. The definitions require that the specified independence hold whenever some subset of variables are held fixed. For instance,

**Definition 2.10:** $X$ and $Y$ are *conditionally additively independent* (CA-independent) given $Z$ ($X, Y, Z$ disjoint, $X \cup Y \cup Z = V$) iff, for any fixed value $\vec{\gamma}$ of $Z$, $X$ and $Y$ are additively independent in the conditional preference structure over $X \cup Y$ given $\vec{\gamma}$.

In this case, we write $CAI(X, Z, Y)$. ∎

**Proposition 2.11:** $X$ and $Y$ are additively independent given $Z$ iff $u_\succeq$ can be written in the form $f(X, Z) + f(Z, Y)$.

## 3   Conditional Additive Independence

Our main new results concern CA-independence and are presented in this section. As mentioned above, the concept of CA-independence has been defined in the literature, but we have found rather little development of the idea. We nevertheless feel that CA-independence is a useful notion for artificial intelligence. In particular, it is not as strong a requirement as additive independence: it is quite feasible that some variables that are not additively independent become additively independent when the values of some other variables are fixed. Furthermore, while it is not as generally applicable as utility independence, utility independence often does not yield a decomposition of the utility function that is as computationally useful. In fact, the decomposition yielded by CA-independence can significantly improve the efficiency of computing expected utility (see Section 3.1).

Our results here are that CA-independence is particularly well suited for graphical modeling. In brief, we show the following. First, every utility function has a *perfect* CA-independence graph: a graph in which vertex separation corresponds exactly to CA-independence. And second, it is possible to read directly from the graph the most general functional form for utility functions satisfying the represented independencies. In the presence of nontrivial independencies, this form typically has a much reduced "dimensionality", making elicitation, representation, and reasoning far easier.

Our first definition defines the type of functional form we are after. We want the utility to be composed from functions with proper subsets of $V$ as arguments—the fewer the arguments the better, as the complexity of specifying a function explicitly (i.e., as a table) is exponential in the number of arguments. Furthermore, as mentioned above, the utility should be a linear combination of these subfunctions.

**Definition 3.1:** Let $Z_1, ..., Z_k$ be (not necessarily disjoint) subsets of $V$. A function $f(V)$ has an additive decomposition over $Z_1, ..., Z_k$ if

$$f = \sum_{i=1}^{k} f_i(Z_i)$$

for some functions $f_i$. ∎

Clearly there is no loss of generality to assume that for no $i, j$ is $Z_i \subseteq Z_j$.

This form of functional decomposition has been used before in the literature [JJD94, DDP88, ST90], but without any justification in terms of notions of utility independence. As Proposition 2.11 shows a utility function can be written in this form only if some collection of CA-independencies hold.

Using this definition, we may rephrase Proposition 2.11 as saying that a utility function satisfies $CAI(X, Z, Y)$ if and only if it has an additive decomposition over $X \cup Z, Z \cup Y$. Although we are ultimately interested in the components *present* in an additive decomposition (in particular, making them as few and as small as possible), proving the main theorems below sometimes requires that we focus on components that are absent instead.

**Definition 3.2:** Let $Z_1, \ldots, Z_k$ be (not necessarily disjoint) subsets of $V$. A function $f(V)$ has an additive decomposition that *avoids* $Z_1, \ldots, Z_k$ if

$$f = \sum_{i=1}^{\ell} f_i(Y_i)$$

for some $f_i$ and some $\ell$ subsets $Y_i \subset V$ such that for no $i, j$ is $Z_i \subseteq Y_j$. ∎

It is easy to verify using Proposition 2.11 that $CAI(X, Z, Y)$ iff there is an additive decomposition avoiding all $\{x, y\}$ such that $x \in X$ and $y \in Y$; we use this in the proofs below.

If $u$ has a decomposition avoiding $X$, and another avoiding $Y$, does it have another decomposition avoiding both? It might seem plausible that there are functions in which an interaction term in either $X$ or $Y$ is necessary, but such that either one of these suffices. However, this is not in fact possible, and the answer to the question above is always yes. The next lemma, which generalizes this claim, will be used in several places subsequently.

**Lemma 3.3:** *If a utility function $u(V)$ has decompositions avoiding each of $X_1, X_2, \ldots, X_k \subset V$ separately then it has a decomposition avoiding them all.*

**Proof:** Omitted. ∎

Now we are in a position to prove our first main result. This says that, for any utility function, there is an undirected graph $G = (V, E)$ (i.e., the nodes are the attributes) such that $CAI(X, Z, Y)$ (for $X \cup Y \cup Z = V$) if and only if $Z$ separates $X$ from $Y$, i.e., every path from a node in $X$ to a node in $Y$ passes through some node in $Z$. In the terminology of [Pea88] such a graph is said to be a perfect map of the independence structure. Pearl and Paz [PP89] have given necessary and sufficient conditions for an independence relation to have such a map (see also [Pea88]), and we can simply apply this result in the proof below. Note that if follows that if two attributes are linked by an edge in a perfect graph for CA-independence, then we always care about the (probabilistic) correlation between their values. If there is no edge then, once all the remaining variables are given fixed values, we do not care whether the endpoint attributes are determined independently or not (i.e., this is irrelevant to our preferences).

**Theorem 3.4:** *The set of CA-dependencies generated by any utility function has a perfect map.*

**Proof:** It is sufficient to simply check each of the five conditions given by Pearl and Paz. The main difficulty is that these conditions concern expressions of the form $I(X, Y, Z)$ in which $X \cup Y \cup Z$ need not equal $V$, but we have not defined CA-independence in this case. So, for the purposes of this proof only, we make the following definition. If $R = V - X - Y - Z \neq \emptyset$, then $CAI(X, Y, Z)$ holds iff there is some partition $R = R_1 \cup R_2$ such that $CAI(X \cup R_1, Z, Y \cup R_2)$. We now simply verify the conditions, which must hold for all sets of disjoint arguments.

**Symmetry:** $CAI(X, Z, Y) \Rightarrow CAI(Y, Z, X)$. This is immediate by the definition.

**Decomposition:** $CAI(X, Z, Y \cup W) \Rightarrow CAI(X, Z, Y) \wedge CAI(X, Z, W)$. This follows from the extended definition give above (because we can choose $R_2$ to contain $W$ or $Y$ as appropriate).

**Intersection:** $CAI(X, Z \cup W, Y) \wedge CAI(X, Z \cup Y, W) \Rightarrow CAI(X, Z, Y \cup W)$. Let $R = V - X - Y - Z - W$. By assumption, there are two partitions of $R$, $R_1, R_2$ and $R'_1, R'_2$ say, such that $CAI(X \cup R_1, Z \cup W, Y \cup R_2)$ and $CAI(X \cup R'_1, Z \cup Y, W \cup R'_2)$ hold. Let $R''_1 = R_1 \cap R'_1$ and $R''_2 = R_2 \cup R'_2$; note that $R''_1 \cup R''_2 = R$. From the above it is easy to verify that, for each $v_1, v_2$ with $v_1 \in X \cup R''_1$ and $v_2 \in Y \cup W \cup R''_2$, there is an additive decomposition avoiding $\{v_1, v_2\}$. By the lemma, there is an additive decomposition avoiding them all at once. Thus $CAI(X \cup R''_1, Z, Y \cup W \cup R''_2)$, i.e., $CAI(X, Z, Y \cup W)$ as required.

**Strong union:** $CAI(X, Z, Y) \Rightarrow CAI(Y, Z \cup W, X)$. This follows easily, because each $v \in W$ is either in $R_1$ or $R_2$ and so, according to the antecedent, the utility can be decomposed so that $v$ appears with $X \cup Z$ or $Y \cup Z$ but not both. The consequent allows it to appear with both, and so is strictly weaker.

**Transitivity:**
$CAI(X, Z, Y) \Rightarrow CAI(Y, Z, w) \vee CAI(w, Z, Y)$ where $w$ is any single variable. By disjointness, we do not consider $w \in Z$. Otherwise, find $R_1$ and $R_2$ such that $CAI(X \cup R_1, Z, Y \cup R_2)$; these must exist by definition. But then $w \in R_1$ or $w \in R_2$, and in either case the result follows immediately using decomposition.

Thus, we can appeal to Pearl and Paz's result to conclude the proof. ∎

It follows from the proof of this theorem that any CA-independence model induced by a utility function is a *graphoid*, in the sense of [PP89]. The book [Pea88] discusses graphoids, and their graphical maps, in considerable detail. Note that an undirected graphical model of the type we consider is also called a Markov network.

Part of the utility of this theorem is that we can represent and reason about CA-independencies graphically, which is often far more natural. However, another benefit is that we can read the functional form of the utility function directly from its corresponding Markov network in the standard fashion (i.e., by identifying the *cliques* of the graph).

**Theorem 3.5:** $G = (V, E)$ is a CA-independence map for a utility function $u$ (i.e., all independencies suggested by vertex separation in the graph hold of $u$) if and only if $u$ has an additive decomposition over the set of maximal cliques of $G$.

**Proof:** First, suppose $u$ has such an additive decomposition, and let $X, Y, Z$ be a partition of $V$ such that $Z$ separates $X$ from $Y$ in $G$. We must show that $CAI(X, Z, Y)$ holds of $u$. But no clique in $G$ can contain an element from both $X$ and $Y$ (otherwise, no separator would exist). Thus, the hypothesized decomposition has no term involving variables from both $X$ and $Y$. The result now follows from Proposition 2.11.

Conversely, suppose $G$ is a CA-independence map of $u$. Let $Y$ be a proper superset of any maximal clique. There exists an additive decomposition over cliques if there is a decomposition avoiding all such $Y$. By Lemma 3.3, it suffices to consider each $Y$ separately. Suppose, for a contradiction, that it is impossible to avoid $Y$. Let $X \subset Y$ be a maximal clique. There must be some $y \in Y$ such that not all members of $X$ are connected to $y$ in $G$. (Otherwise, $X \cup y$ is a larger clique). Let $x \in X$ not be connected to $y$ by an edge. But then $V - \{x, y\}$ separates $x$ and $y$, and so $CAI(\{x\}, V - \{x, y\}, \{y\})$, hence there is a decomposition avoiding $\{x, y\}$, and hence avoiding $Y$. This gives the necessary contradiction. ∎

This means that we can read a suitable function form directly from the graph, by finding cliques. Unless the graph is complete, this gives us a nontrivial decomposition of $u$. By Theorem 3.4, the procedure of finding a graph using CA-independence and then using it this way is capable of revealing all the information inherent in CA-independencies. In a sense, this is quite a strong result. Probabilistic independence does not always admit perfect Markov networks. Thus, while probabilistic independence maps are certainly a useful technique, they do not have the same power as if we were to just reason about independence directly. On the other hand, this contrast is a bit misleading. Although graphical models can capture CA-independence perfectly, the concept of CA-independence itself is somewhat weak.

**Example 3.6:** Consider all utility functions over three variables of the form

$$u(x, y, z) = f(x, y) + f(y, z) + f(x, z).$$

It is easy to verify that there are no independencies (utility or additive, conditional or otherwise) that are common to all such functions. The CA-independence is a complete graph (a triangle) and so does not reflect the fact that $u$ has quite a simple form. ∎

This shows that certain linear functional forms, that entail just the computational and representational advantages we seek, are not revealed by the independence concepts seen so far. Are there other concepts of independence that do not have this weakness? The answer is yes.

**Definition 3.7:** Let $Z_1, ..., Z_k$ be sets of variables *not necessarily disjoint* such that $V = \bigcup_i Z_i$. $Z_1, ..., Z_k$ is *generalized* additively independent (for $\succeq$) if, for any probability distributions $p_1$ and $p_2$ that have the same marginals on $Z_i$ for all $i$, $p_1$ and $p_2$ are indifferent under $\succeq$. ∎

This notion of independence is just like additive independence except that the $Z_i$ do not need to be disjoint. Now it can be shown that Proposition 2.9 still holds, but with generalized additive independence instead of additive independence.

**Proposition 3.8:** $Z_1, ..., Z_k$ are generalized additively independent for $\succeq$ iff $u_\succeq$ can be written as

$$u_\succeq(V) = \sum_{i=1}^{k} f(Z_i)$$

for some functions $f_i$.

In other words, from Definition 3.1, the $Z_i$ are generalized additive independent iff the utility function $u$ has an additive decomposition over them.

This shows that any additive decomposition corresponds exactly to a single assertion of general independence. We have not seen the idea of generalized independence as given above defined explicitly in the literature, or Proposition 3.8 noted. We prove this proposition using the following deceptively powerful result of Fishburn's [Fis82].

**Theorem 3.9:** *[Fishburn] Let $\succeq$ be a preference structure over some collection of states $S' \subset S = \prod_{v \in V} d_v$. We say that some partition $Z_1, ... Z_k$ of $V$ are additively independent over $S'$ if all probability distributions $p_1$ and $p_2$ with support in $S'$, that have the same marginals on $Z_i$ for all $i$, are indifferent under $\succeq$.*

*Then $Z_1, ..., Z_k$ are additively independent over $S'$ iff there exist functions $f_i$ such that*

$$u_\succeq(V) = \sum_{i=1}^{k} f_i(Z_i)$$

*is valid on $S'$.*

Fishburn's theorem basically says that Proposition 2.9 continues to hold over *subsets* of the product space (with the appropriate notions restricted to that subset). The ability to restrict to a subset of the product space, and thus impose fixed interdependencies among the variables, is nontrivial; see [KLST71] for other relevant discussion.

**Proof of Proposition 3.8:** Our proof utilizes a technique suggested by Fishburn in [Fis82]. Let $S = \prod_{v \in V} d_v$, and consider any $Z_i = \{v_1, ..., v_\ell\}$ say. Corresponding to this, we can construct a new variable $z_i$, whose domain is isomorphic to $\prod_{j=1}^{\ell} d_j$. Now consider the space $T = \prod_{j=1}^{k} z_i$. Each $s \in S$ corresponds to an element of $T$ (because $S$ implies a unique value for each of the sets of variables $Z_i$ and thus for $z_i$), and thus $S$ corresponds to a subset $T'$ of $T$. Instead of probability and preference over $S$, we can equivalently consider probability and preference over $T'$. But note that the marginal probability of $Z_i$ in $S$ is equal to the marginal probability over $z_i$ in $T'$. Thus the assumption of Theorem 3.9 is in fact equivalent to the precondition of Proposition 3.8. Hence, the preference structure corresponds to a utility function $u(V) = \sum_{i=1}^{k} f(z_i) \equiv \sum_{i=1}^{k} f(Z_i)$. ∎

Of course, even though any additive decomposition corresponds to a single generalized independence assertion, it is probably unreasonable to try to discover the latter directly. Thus simpler but more accessible concepts such as (plain) CA-independence will remain important.

### 3.1 Computation

It should be clear that an additively decomposable utility function has advantageous computational implications. For instance, if we have no more concise representation of utility than just $u(V)$, we must consider each possible state in $|S|$ individually. This is true no matter how the probability distribution is given to us. But towards the other extreme, if $u$ has an additive decomposition over $v_1, \ldots, v_k$ (i.e., $u(V) = \sum_{i=1}^{n} f_i(v_i)$) then we only need to find $n$ marginal probabilities, because the expected utility of $u$ is the sum of the expected utilities of the $f_i$, by linearity of expectation. Finding these probabilities can be very easy (for instance, linear time given a singly connected Bayesian network; see [Pea88]). This example shows that the advantage of additive utility independence is not simply the reduction of dimensionality. Other decompositions may be as good in this respect, but not offer any clear benefits for computation; for example, consider the product utility functions that can be entailed by mutual utility independence (Propositions 2.5 and 2.6).

To show the possible advantages in somewhat more detail, consider probability distributions given by general Bayesian networks. One of the most popular ways of computing probabilities from a network is to form a join tree; see [LS88, Pea88]. Without going into details, we note that the join tree is a tree of sets of variables. If, for instance, $C = v_1, v_2, \ldots, v_k$ is a node in the join tree then the domain of $C$ is just the product space $\prod_{i=1}^{k} d_{v_i}$. Join trees can be used to maintain the marginal distribution over all nodes $C$; the complexity of this process is determined by the domain sizes of nodes such as $C$ (which can of course be exponential, although in many cases will be of reasonable size).

Suppose, however, that $u$ is decomposable over $Z_1, \ldots, Z_k$ and each $Z_i$ is a subset of some node in the join tree. In this case, expected utility computations can be performed essentially for free, "piggy-backing" on the probability calculations in the probabilistic join tree. Again, this is a consequence of linearity of expectation and the fact that the marginal over $C$ is enough to calculate the expectation of any $f(Z_i)$ with $Z_i \subseteq C$. If this *containment property* (see [DDP88]) property does not hold, we may need to add edges to the Bayesian network or join tree to establish it.

In this case, the extra cost involved in calculating expected utility is a function of the number of edges we need to add. Roughly speaking, the greater the similarity between the Bayesian network (or join tree) and the utility graph, the less extra work will be required to compute expected utility. The technique presented by Jensen et al. [JJD94] for evaluating influence diagrams uses such ideas, as does the somewhat related proposal of Dechter et al. [DDP88] in the context of constraint satisfaction.

Another very relevant work is [ST90], which also uses decomposable utility functions (there called "separable") for evaluating influence diagrams. But this paper deals with both additive and multiplicative decompositions, and so perhaps does not offer as much potential savings as, say, Jensen et al.'s proposal. Finding out whether this is really so would be helped by a more precise analysis of how a utility independence independence structure, and its similarity (or otherwise) to a given probabilistic independence structure, can affect computational efficiency. We hope to address this issue in future work.

## 4 Conclusions and Future Work

A direct extension of this work would be to investigate the possibility and usefulness of graphical models for other relevant concepts of independence. Von Stengel [vS88], utilizing the work of Gorman [Gor68], has shown that a utility function can be graphically represented as a composition tree that captures its utility independencies. The nodes of this tree are subsets of variables, with the root being equal to $V$. Besides this work on utility independence however, there seems to have been little else done in this area.

For instance, are there models using directed graphs (like Bayesian networks) for additive independence? What about generalized additive independence? The case of utility independence also deserves further attention. Gorman's composition tree approach leaves us with a graphical model that is quite distinct from graphical probabilistic models where the nodes represent single variables rather than sets of variables. Hence, it is not clear how such a model can be utilized in conjunction with modern techniques of probability structuring.

It may turn out that the use of graphical models for utility representation only has a limited usefulness. On a broader level, though, we are convinced that decision theory is a critical part of artificial intelligence and thus there should be more work, in various directions, towards more sophisticated utility modeling. Having said this, research in utility modeling need not start from nothing, but can and should draw on the considerable amount of existing ideas and techniques in other disciplines.

We close by discussing one fairly speculative topic for future research. Although we are firm believers in the standard decision theory paradigm, it is surely the case that utility (like, perhaps, probability) is not always a concept that the AI "end-user" should have to deal with directly. Utilities determine the purpose of a decision making process, but it seems at least as common and sometimes more natural to specify this purpose using such concepts as goals, constraints, and so on. If we could give the idea of a "goal" clear and complete semantics in terms of utility functions, we could compile a specification in terms of goals into one in terms of utility (so that decision theory could then be used). But finding such semantics is certainly not a trivial task. We must cope with interacting and even contradictory goals, conditional goals, and more. We should also interpret goals in a natural fashion, which is likely to demand default reasoning of some sort. For instance, given two separate goals $A$ and $B$, it is often reasonable that the conjunction $A \wedge B$ is a desirable thing to achieve as well, unless there is a specific reason to believe otherwise. This is something like a default assumption of independence over goals. Taking this idea literally would involve ideas such as Proposition 2.6, but there appear to be numerous complicating factors including: the rich structure of goals and preference stated in a logical language, which may not match the nice factorization of the state space assumed in Section 2; the implications of making this independence a default, and the interaction with all the other default assumptions that might be necessary; the notion of conditional goals and the various logical issues they raise; and so on. As we have said, some work in this direction exists [Bou94, DW91, DSW91, DW94, TP94], but there remains much to be done.

## References

[Bou94]   C. Boutilier. Towards a logic of qualitative decision theory. In *Principles of Knowledge Representation and Reasoning: Proceedings of the 4'th International Conference (KR'94)*, 1994.

[DDP88]   R. Dechter, A. Dechter, and J. Pearl. Optimization in constraint networks. In R. M. Oliver and J. Q. Smith, editors, *Influence Diagrams Belief Nets and Decision Analysis*, pages 411–425. Wiley, 1988.

[DSW91]   J. Doyle, Y. Shoham, and M. P. Wellman. A logic of relative desire (preliminary report). In *Proc. 6th International Symposium on Methedologies for Intelligent Systems*, pages 16–31, 1991.

[DW91]    J. Doyle and M. P. Wellman. Preferential semantics for goals. In *Proc. 9th National Conference on Artificial Intelligence (AAAI '91)*, pages 698–703, 1991.

[DW92]    J. Doyle and M. P. Wellman. Modular utility representation for decision-theoretic planning. In *Proc. 1st International Conference on Artificial Intelligence Planning Systems (AIPS-92)*, pages 236–242, 1992.

[DW94]    J. Doyle and M. P. Wellman. Representing preferences as *ceteris paribus* comparatives. In *AAAI Spring Symposium on decision-theoretic planning*, pages 69–75, 1994.

[Fis82]   P. C. Fishburn. *The Foundations of Expected Utility*. Reidel, Dordrecht, 1982.

[Fre88]   S. French. *Decision Theory*. Ellis Horwood, Chichester, West Sussex, England, 1988.

[Gor68]   W. M. Gorman. The structure of utility functions. *Review of Economic Studies*, 35:367–390, 1968.

[GS88]    P. Gärdenfors and N. Sahlin, editors. *Decision, Probabilility, and Utility: Selected Readings*. Cambridge University Press, Cambridge, 1988.

[JJD94]   F. Jensen, F. V. Jensen, and S. Dittmer. From influence diagrams to junction trees. In *Proc. Tenth Annual Conference on Uncertainty Artificial Intelligence*, 1994.

[KLST71]  D. H. Krantz, R. D. Luce, P. Suppes, and A. Tversky. *Foundations of Measurement*. Academic Press, New York, 1971.

[KR76]    R. L. Keeney and H. Raiffa. *Decisions with Multiple Objectives: Preferences and Value Tradeoffs*. Wiley and Sons, New York, 1976.

[LS88]    S. L. Lauritzen and D. J. Spiegelhalter. Local computations with probabilities on graphical structures and their application to expert systems. *Journal of the Royal Statistical Society B*, 50(2):240–265, 1988.

[MH69]    J. M. McCarthy and P. J. Hayes. Some philosophical problems from the standpoint of artificial intelligence. In D. Michie, editor, *Machine Intelligence 4*, pages 463–502. Edinburgh University Press, Edinburgh, UK, 1969.

[Pea88]   J. Pearl. *Probabilistic Reasoning in Intelligent Systems*. Morgan Kaufmann, 1988.

[PP89]    J. Pearl and A. Paz. Graphoids: A graph-based logic for reasoning about relevance relations. In B. Du Boulay, editor, *Advances in Artificial Intelligence—II*. North-Holland, New York, 1989.

[Sav54]   L. J. Savage. *The Foundations of Statistics*. Dover, New York, 1954.

[SP90]    G. Shafer and J. Pearl, editors. *Readings in Uncertain Reasoning*. Morgan Kaufmann, San Mateo, CA, 1990.

[ST90]    R. D. Shachter and J. A. Tatman. Dynamic programming and influence diagrams. *IEEE Transactions on Systems, Man, and Cybernetics*, 20(2):365–379, 1990.

[TP94]    S. Tan and J. Pearl. Qualitative decision theory. In *Proc. 12th National Conference on Artificial Intelligence (AAAI '94)*, pages 928–932, 1994.

[vS88]    B. von Stengen. Decomposition of multiattribute expected-utility functions. *Annals of Operation Research*, 16:161–184, 1988.